\documentclass{article}

    \PassOptionsToPackage{numbers, compress}{natbib}
\usepackage[final]{main}

\usepackage[utf8]{inputenc} 
\usepackage[T1]{fontenc}    
\usepackage{hyperref}       
\usepackage{url}            
\usepackage{booktabs}       
\usepackage{amsfonts}       
\usepackage{nicefrac}       
\usepackage{microtype}      
\usepackage{xcolor}         

\usepackage{amsmath}
\usepackage{caption}
\usepackage{subcaption}
\usepackage{bbding}
\usepackage{graphicx}
\usepackage{multirow}
\usepackage{tabu}

\AtEndPreamble{
    \usepackage[capitalize]{cleveref}
    \crefname{section}{Sec.}{Secs.}
    \Crefname{section}{Section}{Sections}
    \Crefname{table}{Table}{Tables}
    \crefname{table}{Tab.}{Tabs.}
}

\usepackage{colortbl}
\definecolor{mygray}{gray}{0.95}
\definecolor{my_green}{RGB}{82,208,80}
\usepackage{enumerate}
\usepackage{makecell}
\definecolor{00red}{RGB}{236,35,35}

\usepackage{arydshln} 
\usepackage{pifont}

\title{Uncertainty-Aware Prototype Semantic Decoupling for Text-Based Person Search in Full Images}

\author{%
    \begin{minipage}{\textwidth}
        \centering
        \textbf{Zengli Luo$^{1}$, Canlong Zhang$^{1,2}\textsuperscript{\Envelope}$, Zhixin Li$^{1,2}$, Zhiwen Wang$^3$,Chunrong Wei$^1$} \\
        {\normalfont
        $^1$Key Lab of Education Blockchain and Intelligent Technology, Ministry of Education, Guangxi Normal University, Guilin 541004, China \\
        $^2$Guangxi Key Lab of Multi-source Information Mining \& Security, Guangxi Normal University, Guilin 541004, China \\
        $^3$School of Electronic Engineering, Guangxi University of Science and Technology, Liuzhou 545006, China \\
        {\small \Envelope~Corresponding Author}}
    \end{minipage}
}

\begin{document}

\maketitle

\begin{abstract}
  Text-based pedestrian search (TBPS) in full images aims to locate a target pedestrian in untrimmed images using natural language descriptions. 
  However, in complex scenes with multiple pedestrians, existing methods are limited by uncertainties in detection and matching, leading to degraded performance.
  To address this, we propose \textbf{\emph{UPD-TBPS}}, a novel framework comprising three modules: 
  Multi-granularity Uncertainty Estimation (MUE), Prototype-based Uncertainty Decoupling (PUD), and Cross-modal Re-identification (ReID).
  MUE conducts multi-granularity  queries to identify potential targets and assigns confidence scores to reduce early-stage uncertainty. 
  PUD leverages visual context decoupling and prototype mining to extract features of the target pedestrian described in the query. 
  It separates and learns pedestrian prototype representations at both the coarse-grained cluster level and the fine-grained individual level, thereby reducing matching uncertainty.
  ReID evaluates candidates with varying confidence levels, improving detection and retrieval accuracy.
  Experiments on CUHK-SYSU-TBPS and PRW-TBPS datasets validate the effectiveness of our framework.
\end{abstract}

\section{Introduction}
Text-based person search in full images is a more practical and scalable solution than conventional person ReID, particularly in real-world scenarios where image queries are unavailable~\cite{MARS}.
However, full-scene scenarios introduce significant challenges, such as cluttered backgrounds, dynamic environments, and pedestrian occlusion~\cite{overview24}.  
These challenges can be summarized as: (1) the difficulty of accurately detecting pedestrians in complex scenes where conventional detectors often fail;  
(2) the uncertainty in identifying the correct individual among multiple detected candidates; and (3) the semantic gap between textual descriptions and visual features~\cite{survey24}.  

\begin{figure}
  \centering
  \includegraphics[width=1.0\linewidth]{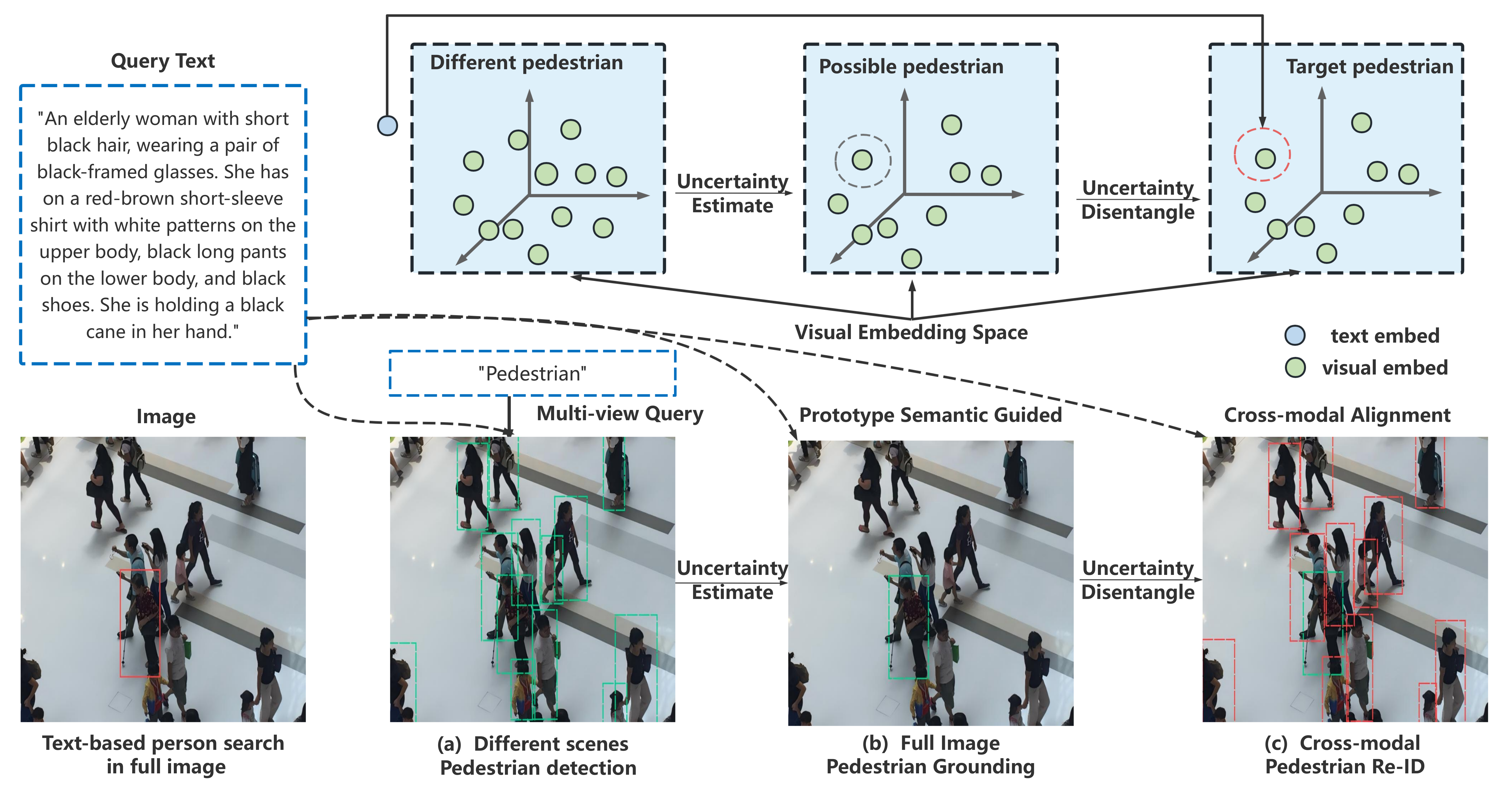}  
  \caption{Illustration of text-based person search in full images task.}
  \label{fig1}
\end{figure}

Despite its real-world relevance, research on text-based person search in full-scene images remains limited~\cite{SDRPN,maca}. Existing methods often suffer from architectural constraints that limit their robustness and accuracy in complex environments and under cross-modal matching conditions.  
To address these issues, we propose a novel framework (see Fig.~\ref{fig1}), which decomposes the text-based person search task into three key sub-tasks:  
(a) pedestrian detection in diverse full-scene environments,  
(b) identification of the target pedestrian within the image, and  
(c) cross-modal re-identification based on textual guidance.  

Motivated by the need to quantify and handle uncertainty throughout this process, we introduce an \textbf{Uncertainty-driven Prediction and Decoupling framework (UPD-TBPS)}. UPD-TBPS estimates and decouples uncertainty to more reliably identify the correct pedestrian among multiple candidates using semantic cues in complex scenes.
\textbf{Pedestrian detection} (Fig.~\ref{fig1}(a)) faces challenges such as viewpoint variation and occlusion, which degrade detection reliability.  
To mitigate this, we introduce the \textbf{Multi-granularity Uncertainty Estimation (MUE)} module, which performs coarse screening and confidence scoring for candidate pedestrians. By estimating detection uncertainty at multiple levels, MUE improves robustness in diverse visual conditions~\cite{uncertaintyAAAI24,uncertainty22}.
\textbf{Pedestrian identification} (Fig.~\ref{fig1}(b)) aims to select the target individual matching the given text from all detected candidates. To address semantic misalignment and intra-modal ambiguity, we design the \textbf{Prototype-based Uncertainty Decoupling (PUD)} module.  
Unlike prior works that treat modality discrepancy~\cite{transCP}, grounding~\cite{transvg}, or semantic prototyping~\cite{prototype24} separately, PUD jointly learns semantic prototypes to serve as both cross-modal anchors and intra-modal regularizers. This improves alignment and stability during matching in cluttered scenes.
\textbf{Cross-modal re-identification} (Fig.~\ref{fig1}(c)) is the final step, which ranks and selects the correct pedestrian guided by the textual description.  
To achieve this, we introduce a \textbf{ReID} module that fuses embeddings from both modalities and incorporates uncertainty cues from MUE and PUD, enabling more reliable ranking and retrieval~\cite{semantic24,semanticQF23}.
Our contributions can be summarized as follows:
\begin{itemize}
  \item We propose \textbf{UPD-TBPS}, a novel framework for text-based person search in full images, which explicitly models and decouples uncertainty across detection and matching stages.  
  \item We design three core modules: \textbf{MUE} for uncertainty-aware pedestrian detection, \textbf{PUD} for prototype-guided identity recognition, and \textbf{ReID} for uncertainty-integrated cross-modal retrieval.  
  \item Extensive experiments on two public benchmarks demonstrate that UPD-TBPS achieves consistent performance improvements over state-of-the-art methods in both accuracy and robustness.
\end{itemize}

\section{Related Work}
\subsection{Person Search}
Person search methods can be broadly categorized into single-stage and two-stage approaches. Single-stage frameworks (e.g., YOLO~\cite{yolo24}) integrate detection and re-identification in an end-to-end pipeline, offering high efficiency but often at the cost of reduced precision.  
In contrast, two-stage methods (e.g., Faster R-CNN~\cite{fastercnn16}) first generate region proposals and then perform feature extraction and identity matching, generally achieving higher accuracy with increased computational overhead.
Recent developments include Anchor-Free detectors~\cite{anchor21}, which eliminate reliance on predefined anchor boxes by directly regressing keypoints or object centers. Meanwhile, Transformer-based methods leverage global attention mechanisms to model long-range dependencies, allowing unified detection and representation learning.
However, full-scene person search remains challenging. Traditional pipelines often struggle to extract semantically relevant regions and suffer from high computational cost due to region proposal networks (RPN) and non-maximum suppression (NMS).
To overcome these limitations, we adopt a separable transformer-based architecture that integrates multi-view query generation and multi-head cross-modal attention. This design enables robust localization of potential pedestrian regions and precise alignment with textual semantics, improving retrieval accuracy while significantly reducing inference complexity.

\subsection{Text-based Person Search}
Text-based person search in full-scene images is particularly challenging due to the inherent cross-modal gap between textual descriptions and visual data, as well as the semantic complexity in cluttered real-world environments.
To address matching ambiguity, \cite{uncertaintyAAAI24} introduced an uncertainty-aware framework that models one-to-many correspondences via uncertainty representations, thereby improving retrieval reliability.  
\cite{transCP} approached the modality discrepancy problem by decoupling visual and textual features and aligning them via prototype-based representations, which improves robustness under semantic variation.
While these methods address specific aspects of the cross-modal problem, they do not jointly tackle uncertainty, visual localization, and semantic alignment in a unified framework.
In contrast, we propose a novel approach that combines uncertainty estimation, prototype-guided feature decoupling, and visual weighting mechanisms within a unified Transformer-based architecture. By jointly addressing cross-modal misalignment and intra-modal ambiguity, our method enhances retrieval precision in complex full-scene scenarios.

\section{Method}
\begin{figure}
    \centering
    \includegraphics[width=1.0\textwidth]{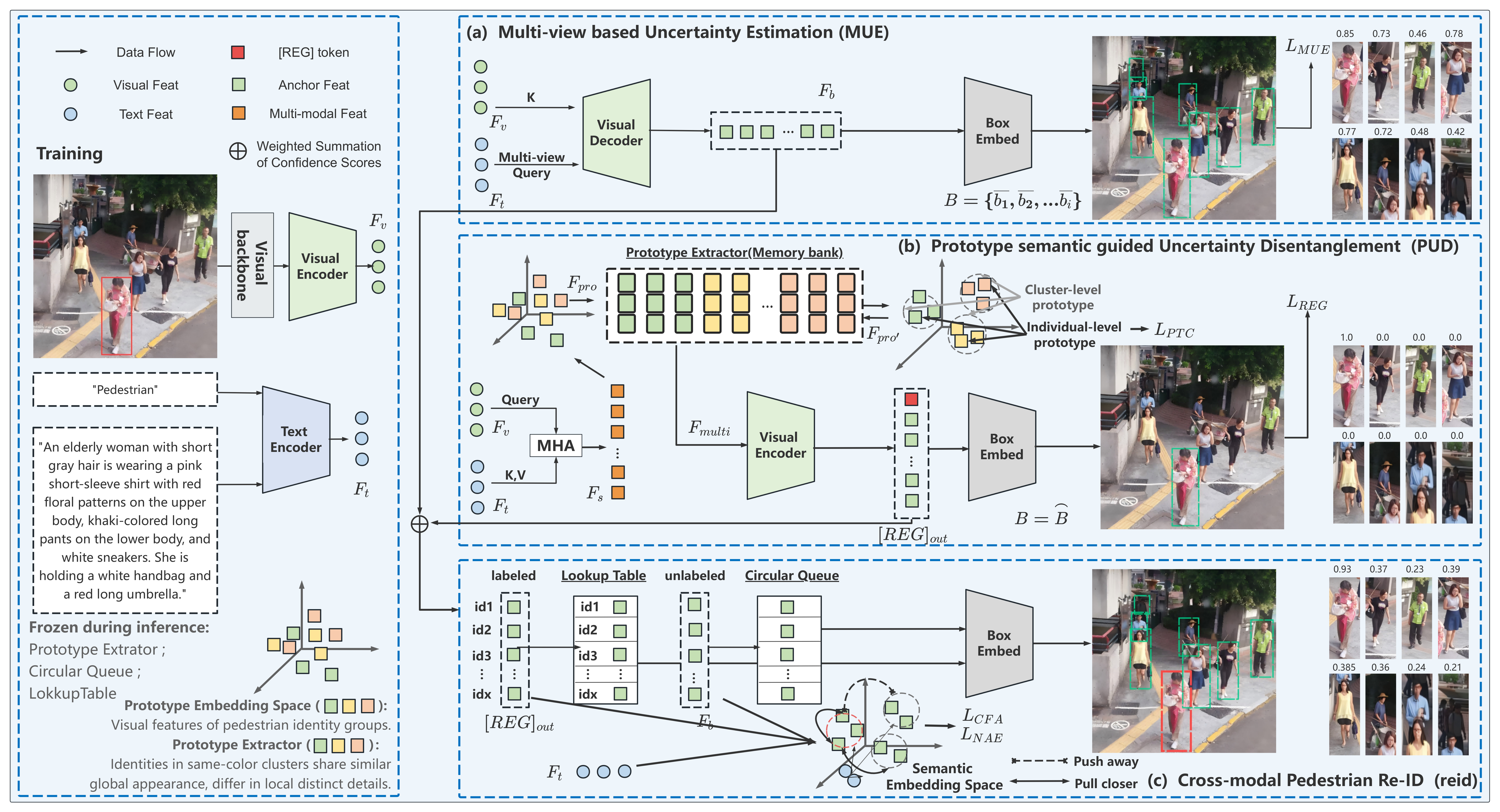}  
    \caption{Overview of the proposed UPD-TBPS framework. It consists of three key components: (a) Multi-Granularity Uncertainty Estimation (MUE), (b) Prototype-Guided Uncertainty Disentanglement (PUD), and (c) Cross-Modal Pedestrian Re-Identification (ReID).}
    \label{fig2}
  \end{figure}
\subsection{Preliminary}
Fig.\ref{fig2} illustrates the overall workflow of the proposed UPD-TBPS framework. 
The primary goal of this framework is to perform person search on uncropped full images by textual descriptions. 
To achieve this, the UPD-TBPS framework progressively narrows the search space and improves retrieval 
precision through three interdependent submodules. 
Specifically, we define $\mathbf{F}_v$ as the visual features of the full images in the training dataset, 
extracted through a trainable collaborative visual encoder, 
and $\mathbf{F}_t$ as the textual features extracted from the 12-layer BERT-based textual encoder\cite{bert} 
for the corresponding textual queries. These features $\mathbf{F}_v$ and $\mathbf{F}_t$ are processed 
sequentially by the three proposed submodules(MUE,PUD,reid), each addressing specific challenges in pedestrian detection. 

\subsection{Multi-granularity based Uncertainty Estimation}
  We adopt a trainable and separable collaborative framework based on a disentangled visual-text encoder-decoder. 
  Different from traditional methods, which are limited to dense queries and unsuitable for pedestrian detection \cite{detrDB1}, this framework integrates multi-granularity textual queries with visual features to search for all potential targets in a full image.
  First, we fuse both global and local visual features $\mathbf{F}_v$ with textual features $\mathbf{F}_t$ obtained from the textual encoder and feed them into a visual decoder. 
  Specifically, the fused visual features serve as the Key, and the textual features serve as the Query.
  As a result, we obtain a set of bounding box features $\mathbf{F}_b$ corresponding to all potential targets related to the queried text. 
  Subsequently, the bounding box Embedding module\cite{detrQF1} transforms these features into a specific set of predicted bounding boxes $\mathbb{B} = \{b_1, b_2, \dots, b_N\}$, 
  where $N$ is the fixed number of predictions made by the visual decoder in a single step. 
  Each predicted bounding box $b_i = \{c_i, l_i\}$ consists of a class prediction $c_i$ and a location prediction $l_i$.
  Then the Hungarian Algorithm is used to optimize the alignment between predicted boxes $\mathbb{B}$ and ground-truth boxes $\hat{\mathbb{B}} = \{\hat{b}_1, \hat{b}_2, \dots, \hat{b}_j\}$. 
  The corresponding loss for this part is given by:
  \begin{equation}
      L_\text{MUE}(\mathbb{B}, \hat{\mathbb{B}}) = \sum_{i=1}^N \left[ -\log p_{\sigma(i)}(c_i) + L_\text{box}(l_i, \hat{l}_{\sigma(i)}) \right],
  \end{equation}
  where $\sigma(i)$ represents the optimal matching index obtained using the Hungarian Algorithm. 
  The term $L_\text{box}$ is the loss for bounding box localization, 
  which is computed as a combination of the L1 loss and the generalized IoU loss, defined as:
  \begin{equation}
  L_\text{box}(l_i, \hat{l}_i) = L_\text{IoU} + \| l_i - \hat{l}_i \|_1.
  \end{equation}
  \subsection{Prototype semantic guided Uncertainty Disentanglement}
  Different from the method based on uncertainty disentanglement of prototype semantics\cite{transCP}, this module learns prototype semantic representations at the instance level and class level, 
  to represent multi-granularity features $\mathbf{F}_\text{multi}$ for the subsequent ReID module. 
  By integrating hard negative mining in fine-grained learning with class-level representation, 
  this module consists of three main steps: (1) generating salient semantic features, (2) prototype mining and aggregation, and (3) collaborative visual-text feature refinement with bounding box regression.  
  First, the visual features $\mathbf{F}_v$ and textual input $\mathbf{F}_t$ are passed through a multi-head attention (MHA) module to generate salient semantic features $\mathbf{F}_s$, expressed as:  
  \begin{equation}
      \mathbf{F}_s = \text{MHA}(\mathbf{F}_v, \mathbf{F}_t).
  \end{equation}
  Next, based on the salient semantic features $\mathbf{F}_s$, a region-specific scaling factor $t$ is computed to augmented the visual features $\mathbf{F}_v$, 
  leading to the refined visual features $\mathbf{F}_\text{pro}$ that focus more on salient targets related to the textual description:
  \begin{equation}
      t = \exp\left( -\frac{(1 - S(\mathbf{F}_v, \mathbf{F}_s))}{2\mu^2} \right),
  \end{equation}
  \begin{equation}
      \mathbf{F}_\text{pro} = \mathbf{F}_v \cdot t,
  \end{equation}
  where $S(\mathbf{F}_v, \mathbf{F}_s)$ measures the similarity between visual and semantic features via projection and dot product, 
  with $\mu$ as a learnable parameter.  
  Multi-head attention is used to suppress background noise and irrelevant text, highlighting target features while reducing distractions.  
  The augmented visual features $\mathbf{F}_\text{pro}$ are then used for prototype mining and aggregation.  
  The prototype set is defined as $\mathbf{P} = \{p_i\}, i = 1, \dots, k$, where $k$ is the number of prototypes, 
  and $p_i$ is the embedding of each prototype.  
  Visual features are assigned to prototypes by finding the nearest one based on Euclidean distance.  
  The index of the nearest prototype is $j = \arg\min \text{dist}(\mathbf{F}_\text{pro}, \mathbf{P})$, 
  and $\mathbf{F}_\text{pro}$ is represented by $\mathbf{P}_j$.Then,cross-modal prototype alignment is achieved using a contrastive learning to ensure consistency between prototypes and textual descriptions.  
  The loss optimizes a symmetric matrix to maximize intra-class similarity and minimize inter-class loss:
  \begin{equation}
      L_\text{proto2t} = -\frac{1}{K} \sum_{i=1}^{K} \log \frac{\exp(\text{sim}(p^m_i, t^m_i)/\tau)}{\sum_{j=1}^{K} \exp(\text{sim}(p^m_i, t^m_j)/\tau)},
  \end{equation}
  \begin{equation}
      L_\text{t2proto} = -\frac{1}{K} \sum_{i=1}^{K} \log \frac{\exp(\text{sim}(p^m_i, t^m_i)/\tau)}{\sum_{j=1}^{K} \exp(\text{sim}(p^m_j, t^m_i)/\tau)},
  \end{equation}
  \begin{equation}
      L_\text{PTC} = \frac{L_\text{proto2t} + L_\text{t2proto}}{2},
  \end{equation}
  where $p^m_i$ and $t^m_i$ represent the embeddings of the $i$-th prototype and its corresponding textual input, $\tau$ is the temperature parameter, 
  and $L_\text{proto2t}$ calculates the loss from the prototype to text, while $L_\text{t2proto}$ calculates the loss from text to the prototype. 
  The similarity $\text{sim}(\cdot, \cdot)$ is computed between embeddings, and the softmax function is used to transform the similarity into a probability distribution.
  
  During training, instance-level features are clustered into prototype centers, forming a prototype bank ($Q_\text{proto}$) updated via a method akin to \cite{StopTrain} for stable representations.  
  Augmented visual features $\mathbf{F}_\text{pro}$ are grouped by prototypes for hierarchical aggregation, producing refined instance-level features.  
  These features, encoding higher-order textual information, are mapped to $Q_\text{proto}$ for efficient inference and used to retrieve multi-granularity candidate features closest to the prototypes.
  
  Subsequently, the refined visual features $\mathbf{F}_\text{pro}$ are fused with the textual features $\mathbf{F}_t$, producing the multi-modal features $\mathbf{F}_\text{multi}$:
  \begin{equation}
      \mathbf{F}_\text{multi} = \tan(\mathbf{F}_\text{pro}) \odot \text{rank}(\mathbf{F}_t),
  \end{equation}
  where $\text{rank}(\mathbf{F}_t)$ represents the rank-ordered textual features, and $\odot$ denotes the Hadamard product.
  
  Finally, the multi-modal features $\mathbf{F}_\text{multi}$ are processed by the visual decoder to extract the features $\text{[REG]}_\text{out}$ for target localization:
  \begin{equation}
      \text{[REG]}_\text{out} = VE\left(\text{[REG]}_\text{in}, \mathbf{F}_\text{multi}, \theta_v\right),
  \end{equation}
  where $\text{[REG]}_\text{in}$ is the input $\text{[REG]}$ token, $VE$ is the cooperative visual decoder, and $\theta_v$ represents the parameters of the decoder.
  
  The extracted features are then fed into the bounding box embedding module to generate the predicted bounding boxes $\hat{B}$:
  \begin{equation}
      \hat{B} = \text{BoxEmbed}\left(\text{[REG]}_\text{out}\right).
  \end{equation}
  
  The regression loss is calculated as:
  \begin{equation}
      L_\text{REG} = L_1(B, \hat{B}) + L_\text{GIoU}(B, \hat{B}),
  \end{equation}
  where $L_1$ represents the $L_1$ loss, and $L_\text{GIoU}$ is the generalized intersection-over-union loss used to optimize the localization accuracy of the bounding boxes.
  
  Therefore, the total loss of this part is defined as:
  \begin{equation}
      L_\text{PUD} = L_\text{PTC} + L_\text{REG}.
  \end{equation}
  \subsection{Cross-modal Pedestrian re-identification}
  In ReID module, feature alignment and discrimination are key to optimizing visual and textual embeddings.  
  We employ stepwise optimization to improve model efficiency and accuracy. Bounding box features $\mathbf{B}$, extracted and processed from the image modality by the cooperative visual decoder, yield $\text{[REG]}_\text{out}$ features. 
  These are combined with textual features $\mathbf{F}_t$ and sent to a Lookup Table and Circular Queue for Norm-Aware Embedding Learning \cite{NAE}.  
  
  Meanwhile, the model improves cross-modal alignment through Spatial Distance Matrix learning\cite{IRRA} at the class level and Image-Text Contrast learning at the instance level.  
  To align image regions with textual descriptions, the model calculates the Cosine similarity $sim(\text{[REG]}_\text{out}, \mathbf{F}_t)$ and converts it into probability distributions using a softmax function with temperature $\rho$.  
  Based on these probabilities,we have the class level loss for the image-to-text direction:
  \begin{equation}
      L_\text{t2i} = KL(p_{ij} \parallel q_{ij}) = \frac{1}{N} \sum_{i=1}^N \sum_{j=1}^N p_{ij} \log \frac{p_{ij}}{q_{ij} + \epsilon},
  \end{equation}
  where $p_{ij}$ is the matching probability between image region $i$ and textual description $j$, $q_{ij}$ is the ground-truth probability, and $\epsilon$ is a small constant to prevent numerical instability. Similarly, the text-to-image direction is calculated in the same way, resulting in:
  \begin{equation}
      L_\text{class-l} = L_\text{i2t} + L_\text{t2i}.
  \end{equation}
  
  The instance-level loss maximizes the similarity of positive samples while minimizing the similarity of negative samples:  
  \begin{align}
      L_\text{ins-l} &= -\log\left(
      \frac{\exp\left(s_{ii}/\epsilon\right)}
      {\exp\left(s_{ii}/\epsilon\right) + \sum_{j \neq i} \exp\left(s_{ij}/\epsilon\right)}
      \right), \label{eq:ITC}
  \end{align}
  where $s_{ij}$ represents the cosine similarity between the $i$-th visual feature $\mathbf{F}_b^i$ and the $j$-th textual feature $\mathbf{F}_t^j$, and $\epsilon$ is a temperature parameter that adjusts the "sharpness" of the similarity distribution.  
  
  Then we have the corresponding cross-modal feature alignment loss:
  \begin{equation}
      L_\text{CFA} = L_\text{class-l} + L_\text{ins-l}.
  \end{equation}
  
  In addition, we adopt the norm-aware embedding\cite{NAE}to enhance feature discrimination. 
  Specifically, this method uses lookup tables and circular queues to unify features, ensuring better consistency across different modalities. 
  The NAE method utilizes $L_2$ normalization and learnable scaling to refine the visual features $\mathbf{F}_b$. 
  The refined textual features $\mathbf{N}_t$ and visual features $\mathbf{F}_b$ are sent to the Lookup Table (LUT) and Circular Queue (CQ) for further processing:
  \begin{equation}
      L_\text{NAE} = \text{OIM}(\mathbf{F}_b, \mathbf{N}_t, \text{LUT}, \text{CQ}),
  \end{equation}
  where LUT stores features with known identities, and CQ is a circular queue for features with unknown identities. Therefore, the loss for the ReID module is defined as:
  \begin{equation}
      L_\text{ReID} = L_\text{CFA} + L_\text{NAE}.
  \end{equation}
  
  \subsection{Training and Inference}
  To balance the contributions of different components during training, we adopt a normalized adaptive loss formulation. Specifically, the total loss is defined as:
  \begin{equation}
      L_\text{total} = \frac{\alpha_1 L_\text{MUE} + \alpha_2 L_\text{PUD} + \alpha_3 L_\text{ReID}}{\alpha_1 + \alpha_2 + \alpha_3},
  \end{equation}
  where \( \alpha_1, \alpha_2, \alpha_3 \) are dynamic weighting factors that adjust the influence of each loss term. 
  This formulation ensures that no single component dominates training and that the model adaptively emphasizes more uncertain or challenging tasks as needed.
  
  During inference, the lookup table and circular queue in the PUD and ReID modules are frozen. 
  The query text is input into the BERT model to extract textual features $\mathbf{Q}_t \in \mathbb{R}^{1 \times 256}$. 
  Similarly, the image is input into the visual backbone and visual decoder to extract visual features $\mathbf{G}_i \in \mathbb{R}^{n \times 256}$, 
  which are then passed to the MUE and PUD modules. 
  The MUE module outputs a set of candidate bounding boxes $\mathbf{B}_\text{MUE}$ and their corresponding confidence scores $\mathbf{C}_\text{MUE}$. 
  The PUD module outputs another set of candidate bounding boxes $\mathbf{B}_\text{PUD}$ and their corresponding confidence scores $\mathbf{C}_\text{PUD}$. 
  The final prediction is obtained by fusing the confidence scores of the candidate boxes from the MUE and PUD modules. 
  Specifically, for a candidate box $b_i$ from the MUE module and a candidate box $b_j$ from the PUD module, if their intersection-over-union (IoU) exceeds a predefined threshold (e.g., $0.5$), 
  the two boxes are considered a match, and their combined score $R_{ij}$ is calculated as:
  \begin{equation}
      R_{ij} = \alpha \cdot c_i + \beta \cdot c_j,
  \end{equation}
  where $c_i$ and $c_j$ are the confidence scores of $b_i$ and $b_j$, respectively. Finally, all matched candidate boxes are ranked by $R_{ij}$, and the box with the highest score is selected as the final prediction.

\section{Experiments}
\subsection{Implementation Details}
We conduct experiments on two benchmark datasets: CUHK-SYSU-TBPS~\cite{SDRPN,CUHK-SYSU}, which contains 11,206 training images with 15,080 boxes (5,532 IDs) and 2,900 query boxes, and PRW-TBPS~\cite{SDRPN,PRW}, with 5,704 training images (14,897 boxes, 483 IDs) and 2,056 queries. Each training/query box in CUHK-SYSU-TBPS is paired with two/one textual descriptions, and vice versa in PRW-TBPS. Following~\cite{SDRPN}, we use mAP and CMC top-K as evaluation metrics with an IoU threshold of 0.5.
Our model adopts ResNet-50~\cite{fastercnn16} as the visual backbone and a DETR-style encoder-decoder as the collaborative visual encoder. Images are resized to 640×640 and trained with SGD for 100 epochs (initial learning rate 0.0001, decayed ×10 after 60 epochs) and a batch size of 32. For the PUD module, we set the prototype size to 2048, embedding dimension to 256, temperature to 0.07, and the circular queue size to 5,000 (known IDs) / 500 (unknown).

\subsection{Comparison with State-of-the-art Methods}
Table~\ref{tab1} compares our approach with existing state-ofthe-art methods on both benchmark datasets.On the CUHK-SYSU-TBPS dataset, our method achieves the best top-1 accuracy of 57.95\%, demonstrating its great performance on this benchmark. 
On the PRW-TBPS dataset, although the mAP and top-1 accuracy are slightly lower than MACA\cite{maca}, our method performs better in terms of top-5 and top-10 accuracy, achieving 53.55\% and 62.67\%, respectively, which are higher than the 52.87\% and 61.93\% of MACA\cite{maca}.
\subsection{Ablation Study}
\textbf{Effectiveness of Each Component.}In Table \ref{tab2},after introducing MUE and PUD module , the model's mAP and top-1 accuracy on the two benchmark datasets improved by 12.51\% and 9.68\% , as well as 7.48\% and 9.68\%, respectively, significantly enhancing the matching performance.  The addition of instance-level prototype learning further improved the model's top-1 accuracy (an increase of 3.44\% on CUHK-SYSU-TBPS and 2.92\% on PRW-TBPS), indicating that instance-level prototype semantic learning helps capture fine-grained matching relationships.

\textbf{Analysis on Different Confidence Levels during Inference.}  
As shown in Table~\ref{tab3}, both the CUHK-SYSU-TBPS and PRW-TBPS datasets achieve the best overall performance when the confidence fusion parameter $\beta$ is set to 0.5.  
This suggests that a balanced contribution from both MUE and PUD modules yields the most effective retrieval results, while relying too heavily on either module (\emph{e.g.}, $\beta = 0.0$ or $\beta = 1.0$) leads to performance degradation.  
Although slight differences exist across datasets, the overall trend remains consistent.  
From the qualitative results in Fig.~\ref{fig7}, we observe that in datasets with smaller or more occluded targets (e.g., PRW-TBPS), assigning relatively more confidence to early-stage proposals (lower $\beta$) can help mitigate missed detections and improve robustness.

\begin{figure}[!t]
  \centering
  \begin{subfigure}[t]{0.48\linewidth}
    \centering
    \includegraphics[width=\linewidth]{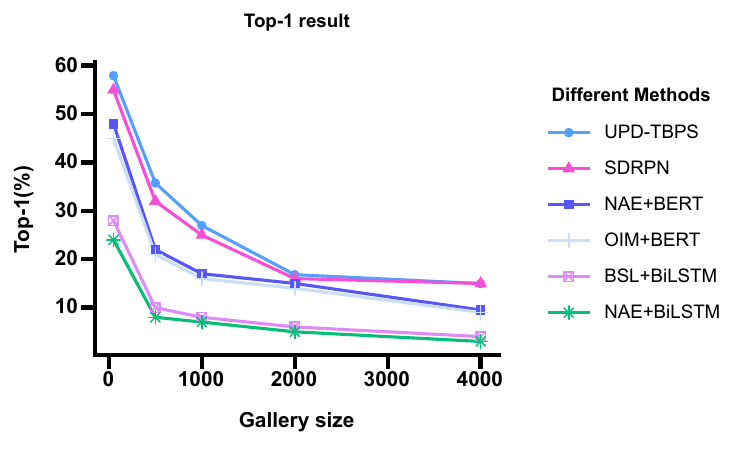}
    \caption{Top-1 result comparison.}
    \label{fig:top1}
  \end{subfigure}
  \hfill
  \begin{subfigure}[t]{0.48\linewidth}
    \centering
    \includegraphics[width=\linewidth]{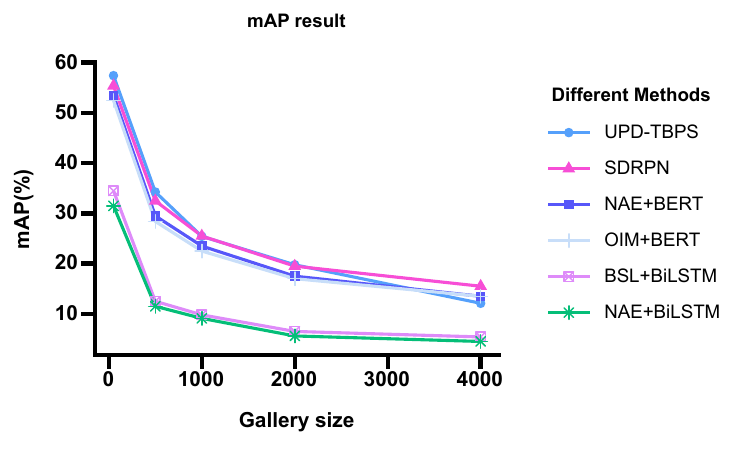}
    \caption{mAP result comparison.}
    \label{fig:map}
  \end{subfigure}
  \vspace{-2mm}
  \caption{Top-1 result and mAP result comparison with different gallery sizes of CUHK-SYSU-TBPS. The left subfigure (\subref{fig:top1}) shows the Top-1 result, while the right subfigure (\subref{fig:map}) shows the mAP result.}
  \label{fig3}
\end{figure}

\textbf{Analysis on Different gallery size on CUHK-SYSU-TBPS.} In Fig.~\ref{fig3}, as the gallery size of CUHK-SYSU-TBPS increases from 50 to 4000, our method consistently outperforms others in both mAP and top-1 metrics. Although performance slightly decreases with the increasing gallery size, our method maintains advantage overall.

\begin{table}[!h]
  \centering
  \caption{Comparisons on CUHK-SYSU-TBPS and PRW-TBPS.}
  \scalebox{0.88}{
  \setlength{\tabcolsep}{3.6pt}
  \begin{tabular}{lllllllll}
  \toprule
  \textbf{Methods} & \multicolumn{4}{c}{\textbf{CUHK-SYSU-TBPS}} & \multicolumn{4}{c}{\textbf{PRW-TBPS}} \\
  \cmidrule(lr){2-5} \cmidrule(lr){6-9}
  & \textbf{mAP} & \textbf{top-1} & \textbf{top-5} & \textbf{top-10} & \textbf{mAP} & \textbf{top-1} & \textbf{top-5} & \textbf{top-10} \\
  \midrule
  OIM\cite{OIM}+BiLSTM & 23.74 & 17.41 & 38.48 & 49.21 & 4.58 & 6.66 & 16.33 & 22.99 \\
  NAE+BiLSTM & 23.48 & 16.62 & 38.45 & 49.66 & 5.20 & 7.54 & 17.21 & 24.11 \\
  BSL+BiLSTM & 26.91 & 20.97 & 42.31 & 52.31 & 3.60 & 6.42 & 15.41 & 22.46 \\
  OIM\cite{OIM}+BERT & 43.39 & 36.59 & 62.03 & 72.66 & 8.52 & 14.44 & 30.68 & 39.77 \\
  NAE+BERT & 45.70 & 39.14 & 64.62 & 74.34 & 9.20 & 14.44 & 31.55 & 39.91 \\
  BSL+BERT & 48.39 & 40.83 & 67.52 & 76.86 & 10.70 & 16.82 & 34.86 & 45.36 \\
  SDRPN\cite{SDRPN} & 50.36 & 49.34 & 74.48 & 82.14 & 11.93 & 21.63 & 42.54 & 52.99 \\
  MACA\cite{maca} & 57.77 & 52.03 & 76.71 & 83.79 & \textbf{18.18} & 33.25 & 52.87 & 61.93 \\
  Ours & \textbf{57.43} & \textbf{57.95} & \textbf{77.36} & \textbf{84.83} & 17.56 & \textbf{37.54} & \textbf{53.55} & \textbf{62.67} \\
  \bottomrule
  \end{tabular}
  }
  \label{tab1}
\end{table}

\begin{table}[!t]
  \centering
  \caption{Ablation Studies with Respect to Model Components on CUHK-SYSU-TBPS and PRW-TBPS. The table shows the performance (mAP, top-1, top-5, and top-10) of different methods on the two datasets. $^*$ indicates results with instance-level prototype semantic learning.}
  \scalebox{0.88}{
  \setlength{\tabcolsep}{3.6pt}
  \begin{tabular}{lllllllll}
    \toprule
    \textbf{Methods} & \multicolumn{4}{c}{\textbf{CUHK-SYSU-TBPS}} & \multicolumn{4}{c}{\textbf{PRW-TBPS}} \\
    \cmidrule(lr){2-5} \cmidrule(lr){6-9}
    & \textbf{mAP} & \textbf{top-1} & \textbf{top-5} & \textbf{top-10} & \textbf{mAP} & \textbf{top-1} & \textbf{top-5} & \textbf{top-10} \\
    \midrule
    RPN+BERT+OIM\cite{OIM} & 41.28 & 36.91 & 64.92 & 71.85 & 9.27 & 12.21 & 27.90 & 37.53 \\
    MUE+OIM\cite{OIM}+BERT & 46.79 & 41.14 & 65.75 & 73.42 & 10.39 & 15.13 & 39.23 & 46.06 \\
    MUE+NAE+BERT & 49.15 & 44.81 & 68.34 & 77.56 & 12.51 & 16.05 & 42.56 & 47.34 \\
    MUE+PUD+OIM\cite{OIM}+BERT & 51.45 & 47.25 & 72.89 & 80.65 & 14.63 & 20.97 & 47.89 & 52.27 \\
    MUE+PUD+NAE+BERT & 53.79 & 48.95 & 73.54 & 81.75 & 16.75 & 21.89 & 49.22 & 53.76 \\
    MUE+PUD$^*$+OIM\cite{OIM}+BERT & 54.95 & 52.39 & 75.63 & 82.89 & 17.04 & 24.81 & 50.55 & 56.02 \\
    Ours & \textbf{57.43} & \textbf{57.95} & \textbf{77.36} & \textbf{84.83} & \textbf{17.56} & \textbf{37.54} & \textbf{53.55} & \textbf{62.67} \\
    \bottomrule
  \end{tabular}}
  \vspace{0.5em}
  \parbox{\linewidth}{\footnotesize{$^*$ with instance-level prototype semantic learning.}}
  \label{tab2}
\end{table}

\begin{table}[!t]
  \centering
  \caption{Performance Comparison of Different Confidence Levels ($\beta$) during Inference on CUHK-SYSU-TBPS and PRW-TBPS. The table presents the performance (mAP, top-1, top-5, and top-10) under varying confidence levels.}
  \scalebox{0.88}{
  \setlength{\tabcolsep}{3.6pt}
  \begin{tabular}{lllllllll}
    \toprule
    \textbf{Confidence Level ($\beta$)} & \multicolumn{4}{c}{\textbf{CUHK-SYSU-TBPS}} & \multicolumn{4}{c}{\textbf{PRW-TBPS}} \\
    \cmidrule(lr){2-5} \cmidrule(lr){6-9}
    & \textbf{mAP} & \textbf{top-1} & \textbf{top-5} & \textbf{top-10} & \textbf{mAP} & \textbf{top-1} & \textbf{top-5} & \textbf{top-10} \\
    \midrule
    $\beta$ = 0.0 & 52.88 & 50.72 & 73.71 & 81.32 & 12.51 & 16.05 & 42.56 & 47.34 \\
    $\beta$ = 0.3 & 56.32 & 53.49 & 76.35 & 83.80 & 17.76 & 26.13 & 51.34 & 56.97 \\
    $\beta$ = 0.5 & \textbf{57.43} & \textbf{57.95} & \textbf{77.36} & \textbf{84.83} & \textbf{17.56} & \textbf{37.54} & \textbf{53.55} & \textbf{62.67} \\
    $\beta$ = 0.8 & 55.22 & 54.71 & 75.16 & 82.03 & 15.76 & 28.47 & 52.58 & 57.07 \\
    $\beta$ = 1.0 & 53.04 & 51.58 & 72.96 & 80.42 & 13.41 & 22.58 & 45.96 & 49.64 \\
    \bottomrule
  \end{tabular}}
  \label{tab3}
\end{table}

\subsection{Visualization}
\begin{figure}[!t]
  \centering
  \includegraphics[width=0.88\linewidth]{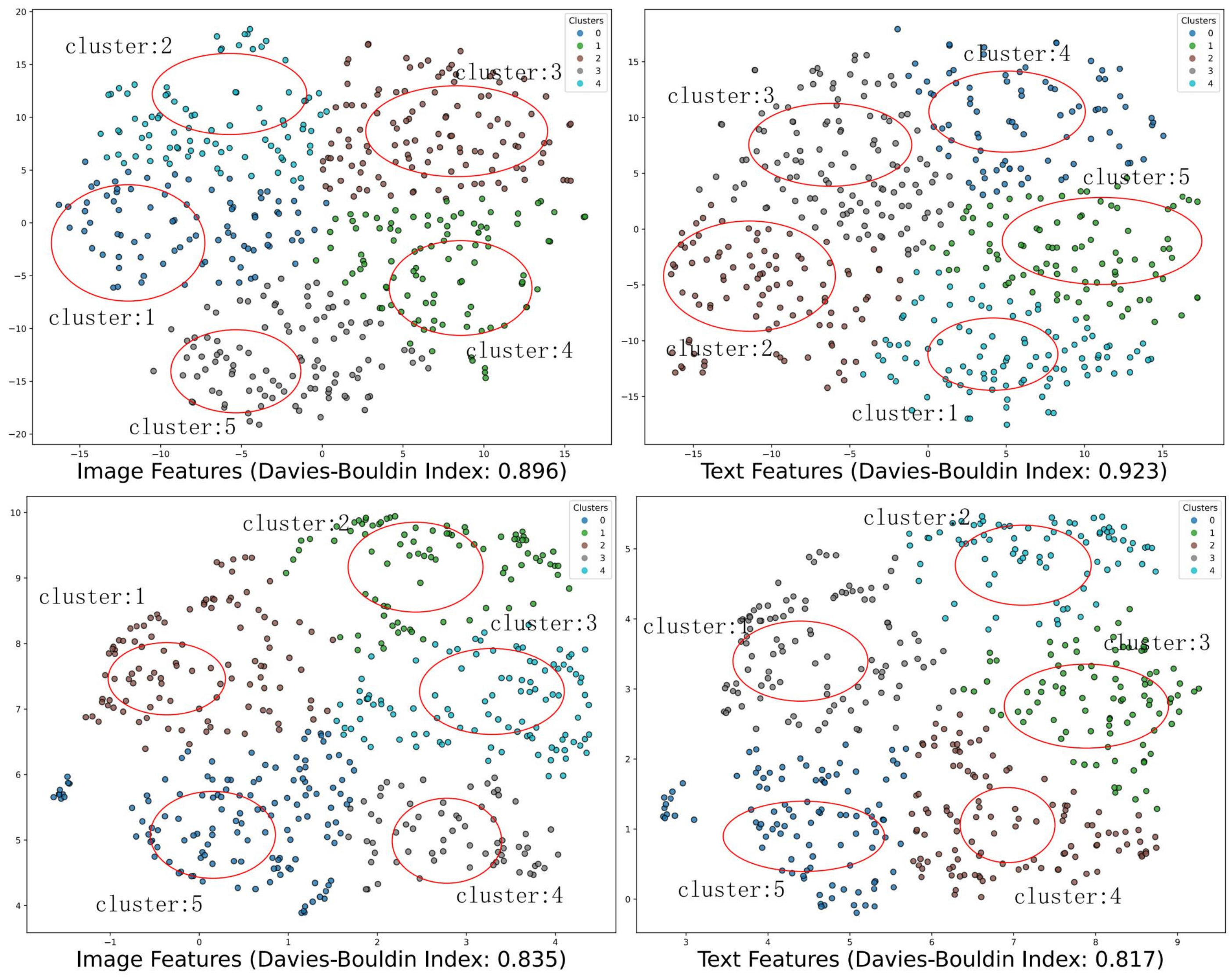}
    \vspace{-2mm}
    \caption{Comparison of image and text features before and after prototype semantic learning at the instance level (top: before, bottom: after).}
    \label{fig4}
\end{figure}

\begin{figure}[!t]
  \centering
  \includegraphics[width=0.88\linewidth]{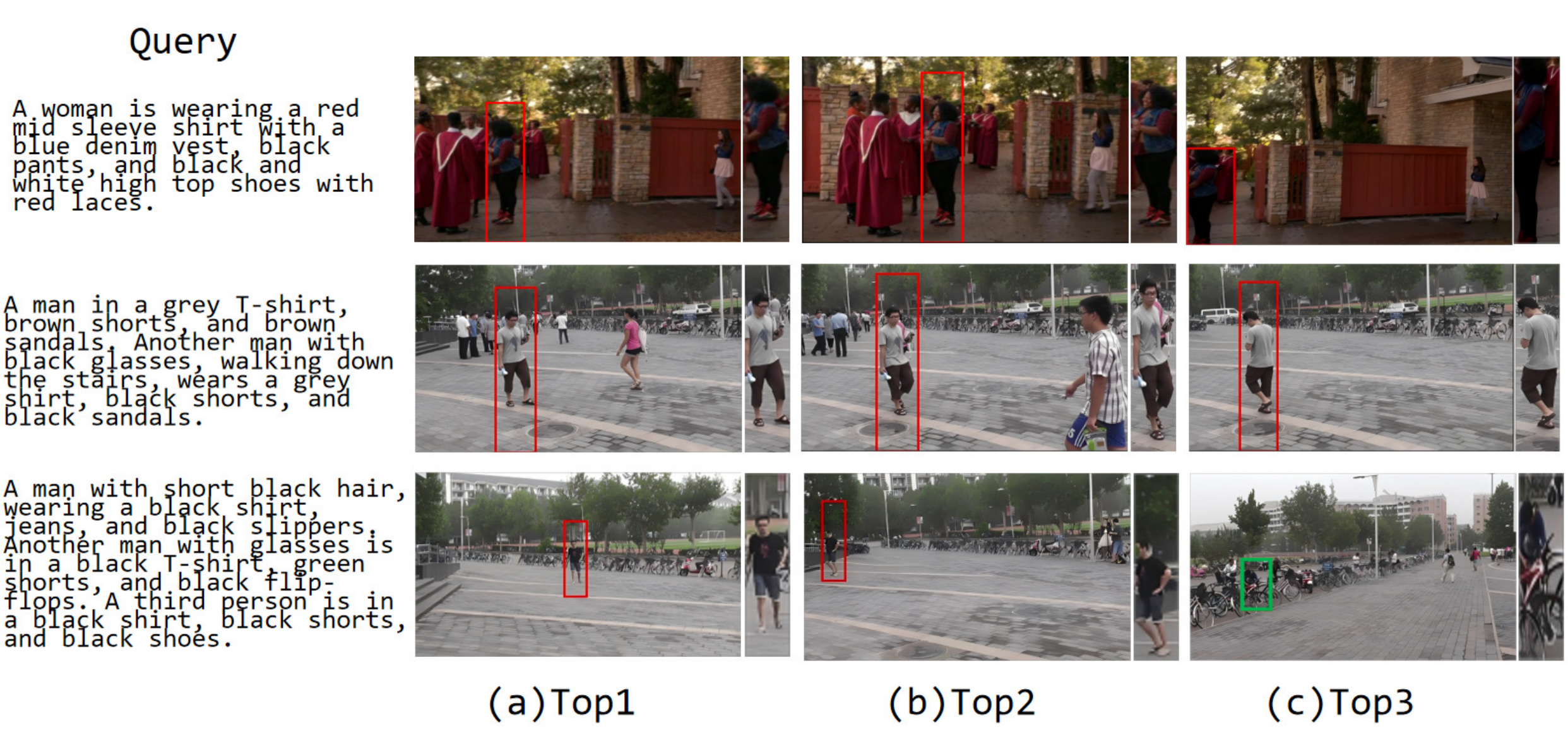}
    \vspace{-2mm}
    \caption{Case studies of text-based person search in full images on CUHK-SYSU-TBPS and PRW-TBPS datasets.}
    \label{fig7}
\end{figure}

Fig.~\ref{fig4} illustrates the clustering results of image and text features before and after prototype semantic learning at the instance level.  
Prior to applying the PUD module, image and text representations exhibit poor alignment, with Davies-Bouldin indices of 0.896 and 0.923, respectively.  
After training, these indices decrease by 6.8\% (image) and 11.5\% (text), indicating improved feature compactness and enhanced cross-modal consistency.
Fig.~\ref{fig7} presents qualitative retrieval examples on the CUHK-SYSU-TBPS and PRW-TBPS datasets.  
Correctly retrieved targets are shown in red bounding boxes, while incorrect ones are highlighted in green.  
UPD-TBPS effectively captures fine-grained semantic cues from textual queries. However, challenges persist in PRW-TBPS due to small target sizes and occlusions, suggesting future directions for model enhancement.

\section{Conclusion}
In this paper, we introduce UPD-TBPS, a text-based person search framework in full images that leverages uncertainty quantification and decoupling. 
By integrating cluster-level and individual-level prototypes through semantic decoupling and prototype learning. Experiments demonstrate its superior robustness and accuracy over state-of-the-art methods on benchmarks. 
Future work will focus on optimizing cross-modal re-identification and addressing occlusion to enhance real-world applicability.

\section{Acknowledgment}
This work is supported by Natural Science Foundation of China (No. 62266009), Guangxi Key Research and Development Program (No. AB25069418), Natural Science Foundation of China (Nos. 62276073, 62466004), Guangxi First-class Undergraduate Course Construction Project (No. 202103) and Innovation Project of Guangxi Graduate Education (XYCSR2024097).

\clearpage

{
\small
\bibliographystyle{unsrt}
\bibliography{main}

\begin{thebibliography}{10}

\bibitem{MARS}
Alex Ergasti, Tomaso Fontanini, Claudio Ferrari, Massimo Bertozzi, and Andrea
  Prati.
\newblock Mars: Paying more attention to visual attributes for text-based
  person search.
\newblock {\em arXiv preprint arXiv:2407.04287}, 2024.

\bibitem{overview24}
Kai Niu, Yanyi Liu, Yuzhou Long, Yan Huang, Liang Wang, and Yanning Zhang.
\newblock An overview of text-based person search: Recent advances and future
  directions.
\newblock {\em IEEE Transactions on Circuits and Systems for Video Technology},
  2024.

\bibitem{survey24}
Prodip~Kumar Sarker, Qingjie Zhao, and Md~Kamal Uddin.
\newblock Transformer-based person re-identification: a comprehensive review.
\newblock {\em IEEE Transactions on Intelligent Vehicles}, 2024.

\bibitem{SDRPN}
Shizhou Zhang, De~Cheng, Wenlong Luo, Yinghui Xing, Duo Long, Hao Li, Kai Niu,
  Guoqiang Liang, and Yanning Zhang.
\newblock Text-based person search in full images via semantic-driven proposal
  generation.
\newblock In {\em Proceedings of the 4th International Workshop on
  Human-centric Multimedia Analysis}, pages 5--14, 2023.

\bibitem{maca}
Liangxu Su, Rong Quan, Zhiyuan Qi, and Jie Qin.
\newblock Maca: Memory-aided coarse-to-fine alignment for text-based person
  search.
\newblock In {\em Proceedings of the 47th International ACM SIGIR Conference on
  Research and Development in Information Retrieval}, pages 2497--2501, 2024.

\bibitem{uncertaintyAAAI24}
Shenshen Li, Chen He, Xing Xu, Fumin Shen, Yang Yang, and Heng~Tao Shen.
\newblock Adaptive uncertainty-based learning for text-based person retrieval.
\newblock In {\em Proceedings of the AAAI Conference on Artificial
  Intelligence}, volume~38, pages 3172--3180, 2024.

\bibitem{uncertainty22}
Myong~Chol Jung, He~Zhao, Joanna Dipnall, Belinda Gabbe, and Lan Du.
\newblock Uncertainty estimation for multi-view data: The power of seeing the
  whole picture.
\newblock {\em Advances in Neural Information Processing Systems},
  35:6517--6530, 2022.

\bibitem{transCP}
Wei Tang, Liang Li, Xuejing Liu, Lu~Jin, Jinhui Tang, and Zechao Li.
\newblock Context disentangling and prototype inheriting for robust visual
  grounding.
\newblock {\em IEEE Transactions on Pattern Analysis and Machine Intelligence},
  2023.

\bibitem{transvg}
Jiajun Deng, Zhengyuan Yang, Tianlang Chen, Wengang Zhou, and Houqiang Li.
\newblock Transvg: End-to-end visual grounding with transformers.
\newblock In {\em Proceedings of the IEEE/CVF International Conference on
  Computer Vision}, pages 1769--1779, 2021.

\bibitem{prototype24}
Shuanglin Yan, Jun Liu, Neng Dong, Liyan Zhang, and Jinhui Tang.
\newblock Prototypical prompting for text-to-image person re-identification.
\newblock In {\em Proceedings of the 32nd ACM International Conference on
  Multimedia}, pages 2331--2340, 2024.

\bibitem{semantic24}
Zhimin Wei, Zhipeng Zhang, Peng Wu, Ji~Wang, Peng Wang, and Yanning Zhang.
\newblock Fine-granularity alignment for text-based person retrieval via
  semantics-centric visual division.
\newblock {\em IEEE Transactions on Circuits and Systems for Video Technology},
  2024.

\bibitem{semanticQF23}
Zheng Wang, Zhenwei Gao, Kangshuai Guo, Yang Yang, Xiaoming Wang, and Heng~Tao
  Shen.
\newblock Multilateral semantic relations modeling for image text retrieval.
\newblock In {\em Proceedings of the IEEE/CVF Conference on Computer Vision and
  Pattern Recognition}, pages 2830--2839, 2023.

\bibitem{yolo24}
Tianheng Cheng, Lin Song, Yixiao Ge, Wenyu Liu, Xinggang Wang, and Ying Shan.
\newblock Yolo-world: Real-time open-vocabulary object detection.
\newblock In {\em Proceedings of the IEEE/CVF Conference on Computer Vision and
  Pattern Recognition}, pages 16901--16911, 2024.

\bibitem{fastercnn16}
Shaoqing Ren, Kaiming He, Ross Girshick, and Jian Sun.
\newblock Faster r-cnn: Towards real-time object detection with region proposal
  networks.
\newblock {\em IEEE transactions on pattern analysis and machine intelligence},
  39(6):1137--1149, 2016.

\bibitem{anchor21}
Yichao Yan, Jinpeng Li, Jie Qin, Song Bai, Shengcai Liao, Li~Liu, Fan Zhu, and
  Ling Shao.
\newblock Anchor-free person search.
\newblock In {\em Proceedings of the IEEE/CVF conference on computer vision and
  pattern recognition}, pages 7690--7699, 2021.

\bibitem{bert}
Jacob Devlin Ming-Wei~Chang Kenton and Lee~Kristina Toutanova.
\newblock Bert: Pre-training of deep bidirectional transformers for language
  understanding.
\newblock In {\em Proceedings of naacL-HLT}, volume~1, page~2. Minneapolis,
  Minnesota, 2019.

\bibitem{detrDB1}
Matthieu Lin, Chuming Li, Xingyuan Bu, Ming Sun, Chen Lin, Junjie Yan, Wanli
  Ouyang, and Zhidong Deng.
\newblock Detr for crowd pedestrian detection.
\newblock {\em arXiv preprint arXiv:2012.06785}, 2020.

\bibitem{detrQF1}
Chunjie Ma, Li~Zhuo, Jiafeng Li, Yutong Zhang, and Jing Zhang.
\newblock Cascade transformer decoder based occluded pedestrian detection with
  dynamic deformable convolution and gaussian projection channel attention
  mechanism.
\newblock {\em IEEE Transactions on Multimedia}, 25:1529--1537, 2023.

\bibitem{StopTrain}
Aaron Van Den~Oord, Oriol Vinyals, et~al.
\newblock Neural discrete representation learning.
\newblock {\em Advances in neural information processing systems}, 30, 2017.

\bibitem{NAE}
Di~Chen, Shanshan Zhang, Jian Yang, and Bernt Schiele.
\newblock Norm-aware embedding for efficient person search.
\newblock In {\em Proceedings of the IEEE/CVF conference on computer vision and
  pattern recognition}, pages 12615--12624, 2020.

\bibitem{IRRA}
Ding Jiang and Mang Ye.
\newblock Cross-modal implicit relation reasoning and aligning for
  text-to-image person retrieval.
\newblock In {\em Proceedings of the IEEE/CVF Conference on Computer Vision and
  Pattern Recognition}, pages 2787--2797, 2023.

\bibitem{CUHK-SYSU}
Shuang Li, Tong Xiao, Hongsheng Li, Bolei Zhou, Dayu Yue, and Xiaogang Wang.
\newblock Person search with natural language description.
\newblock In {\em Proceedings of the IEEE conference on computer vision and
  pattern recognition}, pages 1970--1979, 2017.

\bibitem{PRW}
Liang Zheng, Hengheng Zhang, Shaoyan Sun, Manmohan Chandraker, Yi~Yang, and
  Qi~Tian.
\newblock Person re-identification in the wild.
\newblock In {\em Proceedings of the IEEE conference on computer vision and
  pattern recognition}, pages 1367--1376, 2017.

\bibitem{OIM}
Tong Xiao, Shuang Li, Bochao Wang, Liang Lin, and Xiaogang Wang.
\newblock Joint detection and identification feature learning for person
  search.
\newblock In {\em Proceedings of the IEEE conference on computer vision and
  pattern recognition}, pages 3415--3424, 2017.

\end{thebibliography}
}

\end{document}